\documentclass{article}

\usepackage{spconf}

\usepackage{times}
\usepackage{latexsym}

\usepackage[utf8]{inputenc}
\usepackage{pgfplots}
\DeclareUnicodeCharacter{2212}{-} 
\usepgfplotslibrary{groupplots,dateplot}
\usetikzlibrary{patterns,shapes.arrows}
\pgfplotsset{compat=newest}

\usepackage[T1]{fontenc}
\usepackage[utf8]{inputenc}

\usepackage{microtype}

\usepackage{inconsolata}

\usepackage{graphicx}

\usepackage{multirow} % For multi-row cells
\usepackage{rotating} % For rotating text
\usepackage{wrapfig}
\usepackage{multirow}
\usepackage{amssymb}
\usepackage{amsmath}
\usepackage{amsthm}
\usepackage{bm}
\usepackage{xfrac}
\usepackage{nicefrac}
\usepackage{enumitem,kantlipsum}
\usepackage{algorithm}
\usepackage[noend]{algpseudocode}
\usepackage{verbatim}
\usepackage{todonotes}
\usepackage{xurl}
\usepackage{slantsc}
\usepackage{booktabs}
\usepackage{subcaption}

\algtext*{EndFor}% Remove "end for" text in algorithms
\algtext*{EndIf}% Remove "end if" text in algorithms
\algtext*{EndWhile}% Remove "end if" text in algorithms

\usepackage{verbatim}
\usepackage{mathtools}
\usepackage{subfiles}
\usepackage{enumitem}
\usepackage[mathscr]{eucal}
\usepackage{enumitem,kantlipsum}
\usepackage{xspace}
\usepackage[most]{tcolorbox}
\usepackage{siunitx}
\usepackage{flushend}

\newcommand{\copyrighttext}{%
  \footnotesize \textcopyright 2026 IEEE.  Personal use of this material is permitted.  Permission from IEEE must be obtained for all other uses, in any current or future media, including reprinting/republishing this material for advertising or promotional purposes, creating new collective works, for resale or redistribution to servers or lists, or reuse of any copyrighted component of this work in other works.}
\newcommand{\copyrightnoticefooter}{%
\begin{tikzpicture}[remember picture,overlay]
\node[anchor=south,yshift=10pt] at (current page.south) {\fbox{\parbox{\dimexpr\textwidth-\fboxsep-\fboxrule\relax}{\copyrighttext}}};
\end{tikzpicture}%
}

\usepackage{hyperref}
\hypersetup{
    colorlinks=true,
    linkcolor=blue,
    urlcolor=blue,
}

\DeclareMathOperator*{\argmax}{argmax}
\DeclareMathOperator*{\argmin}{argmin}
\title{DP-LAC: Lightweight Adaptive Clipping for Differentially Private Federated Fine-tuning of Language Models}

\name{Haaris Mehmood\textsuperscript{*}, Jie Xu\textsuperscript{*}, Karthikeyan Saravanan\textsuperscript{*}, Rogier Van Dalen\textsuperscript{\textdagger}, and Mete Ozay\textsuperscript{*}\address{\textsuperscript{*}Samsung R\&D Institute UK (SRUK), \textsuperscript{\textdagger}Samsung AI Centre Cambridge\\
h.mehmood@samsung.com}
}

\begin{document}
\ninept
\maketitle

\copyrightnoticefooter

\begin{abstract}
Federated learning (FL) enables the collaborative training of large‑scale language models (LLMs) across edge devices while keeping user data on‑device. However, FL still exposes sensitive information through client‑provided gradients. Differentially private stochastic gradient descent (DP‑SGD) mitigates this risk by clipping each client’s contribution to a threshold $C$ and adding noise proportional to $C$. Existing adaptive clipping techniques dynamically adjust $C$ but demand tedious hyperparameter tuning, which can erode the privacy budget. In this paper, we introduce DP‑LAC, a method that first estimates an initial clipping threshold within an order of magnitude of the optimum using private histogram estimation, and then adapts this threshold during training without consuming additional privacy budget or introducing new hyperparameters. Empirical results show that DP‑LAC outperforms both state‑of‑the‑art adaptive clipping methods and vanilla DP‑SGD, achieving an average accuracy gain of 6.6\%.

\end{abstract}
\begin{keywords}
Differential Privacy, Federated Learning, Adaptive Clipping, Large Language Model Fine‑tuning
\end{keywords}

\vspace{-0.5em}
\section{Introduction}
Large language models (LLMs) now dominate a wide array of NLP tasks \cite{gpt4, gemini}.  When several stakeholders wish to train such models collaboratively while keeping user data local, federated learning (FL) offers a natural solution: each client trains on its own device and only model updates (pseudo‑gradients) are exchanged with a central server \cite{fedavg, fl, fl_llm}.  Nonetheless, these updates can leak sensitive information, especially for memorizing LLMs \cite{geiping2020inverting, zhang2016understanding, carlini2021extracting, li2022large}.  Differential privacy (DP) \cite{dp_orig} supplies a formal guarantee that protects client contributions against reconstruction attacks \cite{geyer2017differentially, wei2021user}.

In DP, a randomized mechanism \(\mathcal{M}:\mathcal{D}\rightarrow\mathcal{R}\) satisfies \((\varepsilon,\delta)\)‑DP if for every pair of adjacent datasets \(\mathcal{D},\mathcal{D}'\) and every output set $\mathcal{C}\subseteq\mathcal{R}$:
\vspace{-1.0em}
\begin{equation}
    \Pr\!\bigl[\mathcal{M}(\mathcal{D})\in\mathcal{C}\bigr]
\le e^{\varepsilon}\Pr\!\bigl[\mathcal{M}(\mathcal{D}')\in\mathcal{C}\bigr]+\delta
\vspace{-0.15em}
\end{equation}

where \(\varepsilon\) is the privacy budget and \(\delta\) the failure probability \cite{dp_orig}.

\textbf{DP‑FedAvg} \cite{dp_fedavg} implements DP in FL by applying the Gaussian mechanism to each client’s post-training weight difference (pseudo-gradient). The procedure is two‑step:

\begin{enumerate}
    \item Each client clips its pseudo-gradient $\Delta$, to a fixed \(\ell_{2}\)-norm threshold $C$: $\Tilde{\Delta}_k = \Delta_k \cdot \min(1, C / \lVert \Delta_k \rVert_2)$
    \item The clipped update $\Tilde{\Delta}_k$ is perturbed with Gaussian noise\\ \(\mathcal{N}(0,\sigma^{2}I)\), where the standard deviation
    \(\sigma=C\cdot z\) is determined by the noise multiplier \(z\) that
    encodes the client’s privacy parameters \((\varepsilon,\delta)\) and
    participation frequency.
\end{enumerate}

The clipping threshold \(C\) is pivotal: if \(C\) is too large, the added noise dominates the signal; if it is too small, legitimate gradient directions are severely distorted, introducing bias.  Consequently, selecting \(C\) involves a delicate bias–variance trade‑off \cite{adaptive_clipping_du,adaptive_clipping_bu,understanding_clipping}.  In practice, researchers set \(C\) to a percentile (e.g., the 95th) of the distribution of client gradient norms \cite{adaptive_clipping_andrew}, yet the optimal percentile is highly sensitive to model architecture, dataset, and other hyperparameters.  

These observations expose three main limitations of current approaches: (i) manual tuning is time‑consuming and privacy‑costly because each tuning attempt consumes part of the privacy budget; (ii) online clipping schedules introduce additional hyperparameters (e.g., learning rates for the clipping schedule) that must be calibrated, further complicating deployment; and (iii) a fixed \(C\) set at training start yields sub‑optimal robustness, as any shift in data distribution or model dynamics can render the chosen threshold ineffective.

\begin{figure}[t]
    \centering
    \includegraphics[width=1.0\columnwidth]{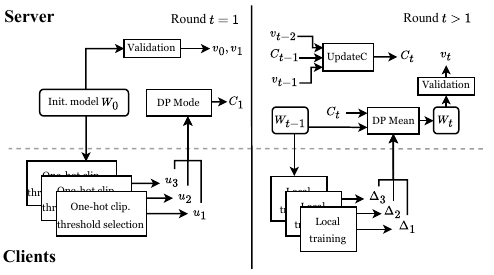}
    \vspace{-2em}
    \caption{Overview of our adaptive clipping method, DP-LAC. $u_i$ is a one-hot encoding vector and $\Delta_i$ the post-training weight difference. $v_t$ is the server-side validation loss and $C_t$ the clipping threshold at round $t$. More details of our method can be found in Algorithm \ref{alg_dplac}.}
    \label{fig:dp-lac-overview}
\vspace{-1.5em}
\end{figure}

Towards this, in this paper, we present \textbf{DP‑LAC} (Differentially Private Federated Fine‑tuning with Lightweight Adaptive Clipping), a method that automatically adapts the clipping threshold C during LLM fine‑tuning under DP‑FL, without adding any extra hyperparameters. In a realistic setup with 1,000 clients and accounting for the privacy cost of hyperparameter tuning, \textbf{DP‑LAC outperforms existing adaptive clipping methods by an average of 6.6\%} with \textbf{5-15x faster hyperparameter grid-search times} \cite{adaptive_clipping_du, adaptive_clipping_bu}. Figure \ref{fig:dp-lac-overview} presents an overview of our method.

\vspace{-0.5em}
\section{Background}
\vspace{-0.5em}
\subsection{Differentially Private Federated Learning (DP-FL)}
\label{sec:dpfl}

\vspace{-0.5em}
Federated learning trains a model from data held by distributed clients while keeping the raw data on each device.  In each communication round \(t\), the server broadcasts the current global model \(\bm{W}_t\) to a randomly selected subset of clients \(\mathcal{K}_t\).  Every client \(k\in\mathcal{K}_t\) performs local training, yielding a locally updated model \(\bm{W}_k\), and sends the model update \(\Delta_k=\bm{W}_k-\bm{W}_t\) back to the server.  The server then aggregates these updates to produce the global model for the next round.

Differential privacy can protect FL at the sample level or at the user level.  In this paper we focus on user‑level DP, introduced by \cite{dp_fedavg}.  To obtain \((\varepsilon,\delta)\)-user‑level DP, the \(\ell_2\)-norm of each client's update \(\Delta_k\) is clipped to a threshold \(C\) and then a Gaussian noise vector \(\mathcal{N}(0,\sigma^2\mathbf{I})\) is added to the sum of updates, where \(\sigma^2=C^2z^2\) and \(z\) is the noise multiplier. This two–stage process is the core of DP‑FedAvg, the defacto algorithm for DP-FL \cite{dp_fedavg}.

\vspace{-0.70em}
\subsection{Adaptive Clipping for DP-FL}

\vspace{-0.60em}
Previous adaptive‑clipping methods \cite{adaptive_clipping_andrew, adaptive_clipping_du, adaptive_clipping_bu, adaptive_clipping_qiu} dynamically adjust the clipping threshold to improve DP optimization.  However, they introduce additional hyperparameters that must be tuned, incurring significant computational effort and consuming part of the privacy budget \cite{dp_hpo_liu, dp_hpo_papernot}. In \cite{dp_hpo_liu} (Theorem 3.5), an algorithm that tunes its hyperparameters achieves at most \((3\varepsilon,0)\)-DP if the underlying algorithm satisfies \((\varepsilon,0)\)-DP.  A concurrent study \cite{bu2025towards} proposes an automatic learning‑rate selection by fitting a quadratic private‑loss function every \(K\) rounds; this approach, however, requires privatising three losses per cycle, introduces a hidden hyperparameter \(\gamma\), and is designed for central DP‑SGD rather than DP‑FL.

\vspace{-0.5em}
\section{Methodology}
\begin{figure}[!t]
\vspace*{-5pt}
\begin{algorithm}[H]
\caption{DP-LAC}
\small
\begin{algorithmic}[1]
\Function{\textsc{ServerLoop}}{}
\State \textbf{parameters: } $T$ (num. rounds), $\mathcal{K}$ (clients), $q\in (0,1]$ (sampling rate),
$\varepsilon$ (privacy budget),
$\delta$ (failure prob.), $E$ (local epochs), $\beta$ (batch size), $\eta$ (local learning rate), $F$ (loss function), $\bm{s} \in \mathbb{R}^d$ ($d$ clip. thresholds),  $\bm{m} \in \mathbb{R}^e$ ($e$ multiplier choices), $\bm{W}_0$ (pretrained model weights), $\mathcal{D}_{val}$ (server validation set)
\State $z \leftarrow$ {\textsc{GetNoiseMult}}$(T, q, \varepsilon, \delta)$ \Comment{Noise multiplier}
\State $v_0 \leftarrow$ $F(\bm{W}_0, \mathcal{D}_{val})$ \Comment{Server validation loss}
\For{$t \gets 1, T$}
\State Sample $\mathcal{K}_t \subseteq \mathcal{K}$ uniformly at random (probability $q$)
\If{$t = 1$}
\State $ C_1 \leftarrow$ {\textsc{InitC}}$(\mathcal{K}_t, \bm{W}_0)$
\State $\bm{W}_1, v_1 \leftarrow \bm{W}_0, v_0$
\Else
\State $ C_t \leftarrow \textsc{UpdateC} \big( C_{t-1}, v_{t-1}, v_{t-2})$
\State  $\bm{W}_{t} \leftarrow \textsc{UpdateW} \big( \mathcal{K}_t, C_t, \bm{W}_{t-1}, z\big)$
\State $ v_t \leftarrow F(\bm{W}_{t}, \mathcal{D}_{val})$
\EndIf
\EndFor
\EndFunction

\Function{\textsc{InitC}}{$\mathcal{K}', \bm{W}, \bm{m}$}
\For{each client $k \in \mathcal{K}'$}
\State $\bm{u}_k \gets \textsc{GetHistC}(k, \bm{s}, \bm{m}, F)$
\EndFor
\State $\Tilde{\bm{u}} = \frac{1}{|\mathcal{K}'|}\big[\sum_{i=1}^{|\mathcal{K}'|}\bm{u}_{i} + \mathcal{N}(0, z^2 \mathbf I)\big]$ \Comment{Private histogram}
\State $\varphi = \argmax_{1 \le i \le d}\Tilde{\bm{u}}^{(i)}$; $C_{\textit{hist}} = \bm{s}^{(\varphi)}$ \Comment{Estimate $C$}
\State \textbf{return} $C_{\textit{hist}}$
\EndFunction

\Function{\textsc{GetHistC}}{$k, \bm{s}, \bm{m}, F$}
\State \textbf{parameters: }$\mathcal{D}_{k}$ (user data), $\bm{u} \gets \bm{0} \in \mathbb{R}^d$, $\bm{l} \gets \bm{0} \in \mathbb{R}^e$
\State $\Delta_k = $ \textsc{UserUpdate}$(k, \bm{W})$; $l_k = F(\Delta_k+\bm{W}, \mathcal{D}_k)$
\State $C_k = \lVert \Delta_k \rVert_2$;
\For{$i \gets 1, e$}
\State $\Tilde{\Delta}_k^{(i)} = \Delta_k  \min\big(1, \frac{\bm{m}^{(i)} C_k}{\lVert \Delta_k \rVert_2}\big)$; $ \sigma = z \bm{m}^{(i)} C_k$
\State $\Bar{\Delta}_k^{(i)} = \Tilde{\Delta}_k^{(i)} + \mathcal{N}(0, \frac{\sigma^2}{\lvert \mathcal{K} \rvert} \mathbf I)$ \Comment{Simulate noise}
\State $\bm{l}^{(i)} = F(\bm{W}+\Bar{\Delta}_k^{(i)}, \mathcal{D}_k)$ \Comment{Loss w/ noisy update}
\EndFor
\State $\pi = \argmin_{1 \le i \le e}(\lvert \bm{l}^{(i)} - l_k \rvert)$
\State $\omega = \argmin_{1 \le j \le d}(\lvert \mathbf{s}^{(j)} - \mathbf{m}^{(\pi)} C_k \rvert)$; $\bm{u}^{(\omega)} = 1$
\State  \textbf{return} $\bm{u}$
\EndFunction

\Function{\textsc{UpdateC}}{$C', v', v''$}
\State \textbf{return} $C'\cdot\min(1, \frac{v'}{v''})$ \Comment{Equation \ref{c-update-rule}}
\EndFunction

\Function{\textsc{UpdateW}}{$\mathcal{K}', C', \bm{W}, z$}
\For{each client $k \in \mathcal{K}'$}
\State $\Delta_k = $ {\textsc{UserUpdate}}$(k, \bm{W})$
\State $\Tilde{\Delta}_k = \Delta_k \cdot \min\big(1, \frac{C'}{\lVert \Delta_k \rVert_2}\big)$; $ \sigma = z \cdot C' $
\EndFor
\State $\bm{W}^{\prime} \leftarrow \bm{W} + \frac1{\lvert \mathcal{K}' \rvert} \big[\mathcal{N}(0, \sigma^2 \mathbf I) + \sum_{k \in \mathcal{K}'} \Tilde{\Delta}_k\big] $
\State \textbf{return} $\bm{W}^{\prime}$
\EndFunction

\Function{\textsc{UserUpdate}}{$k, \bm{W}$}
\State \textbf{parameters: } $\mathcal{D}_{k}$ (user data)
\State $\bm{W}_{k} \leftarrow \bm{W}$
\For{$i \gets 1, E$}
\State $\mathcal{B} \leftarrow$ (split $\mathcal{D}_{k}$ into batches of size $\beta$)
\For{batch $b \in \mathcal{B}$}
\State $\bm{W}_{k} \leftarrow \bm{W}_{k} - \eta \triangledown F(\bm{W}_{k}, b)$
\EndFor
\EndFor
\State \textbf{return} $\bm{W}_k - \bm{W}$ \Comment{Pseudo-gradient}
\EndFunction
\end{algorithmic}
\label{alg_dplac}
\end{algorithm}
\vspace*{-20pt}
\end{figure}

\vspace{-0.7em}
\subsection{Problem formulation}
\vspace{-0.45em}
Fine‑tuning a large‑language‑model (LLM) typically involves updating model weights $\bm{W}$ by minimizing a non‑convex loss
\(F(\bm{W})\) for \(T\) iterations of stochastic gradient descent (SGD). In standard non‑convex SGD, the expected average squared gradient norm converges
to a finite value (often zero) \cite{bottou2018optimization}.

Under DP-FL, the clipping bound \(C\) is a critical hyperparameter balancing privacy and utility. When DP noise $z$ is
injected, a fixed \(C\) becomes increasingly detrimental as the average gradient
magnitude decays, resulting in the noise term \(zC\) to dominate the signal and
potentially destabilise the loss landscape
\cite{dp_fedavg,amin2019bounding,adaptive_clipping_andrew}. A line of work to ensure stable loss landscapes involves adaptively adjusting \(C\) as the training progresses \cite{adaptive_clipping_andrew, adaptive_clipping_du, adaptive_clipping_qiu}, i.e. `adaptive-clipping'.

Existing adaptive‑clipping methods lower \(C\) during training by tuning several predefined hyperparameters (see Table~\ref{tab:grid-search});
this hyperparameter search consumes a non‑negligible portion of the privacy
budget \cite{dp_hpo_liu,dp_hpo_papernot} and takes longer to achieve optimal performance. We seek an online strategy that sets and then continually refines \(C\)
during training while (i) requiring no additional private client statistics and
(ii) introducing no extra hyperparameters.

Instead of relying on client‑side statistics, we observe that a decrease in
training loss naturally implies a reduction in the expected average gradient
norm.  Since the server cannot access private client losses without expending
additional privacy budget, we monitor a public validation set
\(\mathcal D_{\text{val}}\)---a standard practice in federated learning for
continual performance assessment under non‑IID data
\cite{fl}.

Our adaptation rule is as follows: at communication round \(t\), evaluate the validation
loss \(v_{t}=F(\bm{W_t};\mathcal D_{\text{val}})\); if \(v_{t-1}\) decreases relative
to the previous evaluation \(v_{t-2}\), multiply \(C\) by the relative rate of change in $v$.  When no public validation data is available, we split the total
privacy budget---2/3 for standard DP‑FedAvg training and 1/3 for privately estimating client training losses---and estimate the initial loss similarly to $C_{\textit{hist}}$.  In either scenarios, no additional private statistics
are requested from clients and no extra hyperparameters are introduced, satisfying the two desiderata.

\vspace{-0.75em}
\subsection{Lightweight Adaptive Clipping (DP‑LAC)}
\vspace{-0.15em}
\textbf{Online update of the clipping bound:} The only hyperparameter that changes during training is the clipping threshold
\(C\).  To keep the number of knobs minimal we let the server update \(C\) once per communication round using only its own validation loss $v_{t} = F(\bm{W}_{t};\mathcal D_{\text{val}})$:
\vspace{-0.75em}
\begin{equation}
C_{t} = \min(1,v_{t-1}/v_{t-2}) C_{t-1}
\label{c-update-rule}
\end{equation}

\vspace{-0.1em}
Because this update requires previous two successive loss values, we initialise
\(C_{1}=C_{0}\).  The rule automatically shrinks \(C\), matching the intuition that gradients become smaller as the model
converges, while the noise term \(zC\) would otherwise dominate.

\textbf{Choosing the initial clipping bound:} A sensible starting point \(C_{0}\) is essential: a value that is too large
yields excessive noise; a value that is too small introduces severe bias.
During the first round we let each client estimate a locally
optimal clipping norm and report it privately.  With the global model
\(\bm{W}_{t-1}\) broadcast by the server, each client trains on its own data to produce $\bm{W}_t$
and then produces several \emph{noisy variants} \(\{\tilde{\bm
W}_{t}^{(i)}\}\).  The variants are produced by first computing the local
pseudo‑gradient \(\Delta = \bm{W}_{t}-\bm{W}_{t-1}\) and its
\(\ell_{2}\) norm $C_{\textit{init}}$.  The client then selects a small set of \emph{candidate
clipping values} based on $C_{\textit{init}}$, for example \(\{0.25C_{\textit{init}},\,0.5C_{\textit{init}},\,
C_{\textit{init}}\}\).  For each candidate
\(C^{(i)}\), the pseudo‑gradient is clipped, Gaussian noise with variance
\((C^{(i)}z)^{2}\) is added, and the resulting noisy weight update
\(\tilde{\bm{W}}_{t}^{(i)}=\bm{W}_{t}+\tilde{\Delta}^{(i)}\) is formed.
The client then evaluates the loss \(F(\tilde{\bm{W}}_{t}^{(i)};\mathcal
D_{k})\) on its local data for every candidate and chooses the index that is closest to $F(\bm{W}_{t};\mathcal{D}_{k})$.  Only a one‑hot vector indicating that chosen index is
returned; no raw gradients, weights or losses are transmitted.

Because a one‑hot vector has sensitivity of 1, the server can aggregate the
responses using the Gaussian mechanism with clipping threshold set to 1.
The differentially private histogram obtained in this way estimates the
distribution of locally optimal clipping norms across clients; the mode of this histogram is taken as the initial clipping threshold \(C_{0}\). This method follows recent work in federated analytics that privately estimates user statistics \cite{dp-hist-est} and guarantees the same privacy protection
as private SGD \cite{dp_sgd}.

\textbf{Client loss based adaptation (DP-CLAC)}: If a public validation set is unavailable, the server can split the total privacy budget as $2/3$ to privatize the noisy weight updates and $1/3$ to privatize local losses, other splits can also be used. This results in two different noise multipliers so overall the algorithm guarantees the same privacy. During the first round, a similar private histogram of initial loss values is transmitted by clients. For subsequent rounds, each client additionally uploads a single scalar loss value \(l_{k}\).  The server aggregates these
values privately using the Gaussian mechanism with the clipping threshold equal
to the noisy mean of the previous round’s losses. The resulting average
loss provides a signal that can be fed to the same validation‑loss update
as Equation \ref{c-update-rule}.  Because only one scalar per client is transmitted, the
communication overhead is negligible.

\textbf{Full algorithm:} Algorithm \ref{alg_dplac} presents the complete DP‑LAC procedure. \textsc{GetNoiseMult}$(T, q, \varepsilon, \delta)$ computes the noise multiplier \(z\) satisfying \((\varepsilon,\delta)\)-DP via the Gaussian mechanism, given client sampling rate $q$ and total number of FL rounds $T$. It uses a moments accountant \cite{dp_sgd} with Rényi differential privacy (RDP) \cite{rdp} and privacy amplification via subsampling  \cite{privacy_amplification_via_subsampling_1, privacy_amplification_via_subsampling_2} for tighter bounds. An established conversion method in \cite{rdp_accountant} is used to convert $(\varepsilon, \delta)$-DP to $(\alpha, \varepsilon')$-RDP. We use the same moments accountant for all baselines.

%--------------------------------------------------------

\vspace{-0.5em}
\section{Experimental Setup}
\vspace{-0.5em}
\subsection{Hyperparameter regimes}

\noindent\textbf{Default hyperparameters (Def.\ HP):} Every baseline receives the full privacy budget \(\varepsilon\) as specified in the paper.  We perform a single run for each method that reports default settings.  For consistency all methods are started with the same initial clipping threshold \(C=8.0\).

\noindent\textbf{DP hyperparameter optimization (DP\ HPO):} Each baseline’s DP‑related hyperparameters (decay rate, quantile, etc.) are tuned carefully for each privacy budget \(\varepsilon\) and dataset. Following the DP HPO approach of \cite{dp_hpo_liu}, we adjust the privacy budget for each baseline method to \(\varepsilon/3\) so that the total budget consumed post HPO is still approximately $\varepsilon$. Our method, DP‑LAC, does not require any tuning of DP related hyperparameters; consequently it uses the full budget \(\varepsilon\).

\noindent\textbf{Additional experiments:} To illustrate the breadth of our approach we report the wall‑clock cost of hyperparameter tuning, the accuracy of the privately estimated clipping bound \(C_{\text{hist}}\) against the ideal value \(C_{*}\), the performance on a natural‑language generation task, and the robustness of DP‑LAC to changes in LoRA rank.

\vspace{-0.60em}
\subsection{Training details}
\vspace{-0.15em}
\noindent\textbf{Datasets and models:} We evaluate on the GLUE benchmarks (SST‑2, QNLI, MNLI) using the pre‑trained TinyLlama‑1B \cite{tinyllama}.  To test scalability we also fine‑tune the larger Qwen3‑4B model for summarisation on the SAMSum dataset \cite{qwen3}.

\noindent\textbf{Privacy setting:} Three privacy regimes are considered: strong (\(\varepsilon=2\)), medium (\(\varepsilon=4\)), and high (\(\varepsilon=8\)).  A constant \(\delta=10^{-5}\) is used for all experiments.  The noise multiplier \(z\) is derived from the moments accountant for user‑level DP \cite{dp_sgd} exactly as in \cite{dp_fedavg}.

\noindent\textbf{Federated data split:}  For GLUE, we split the training set into \(N=1000\) clients by sampling classes proportionally from a Dirichlet distribution with concentration \(\alpha=1.0\).  For SAMSum, an i.i.d. split over the same number of clients is performed.  Training proceeds for 200 communication rounds on GLUE and 25 rounds  for SAMSum, with one local epoch per round.

\noindent\textbf{Low‑rank adaptation:}  
LoRA \cite{lora, ffa_lora} is applied to the query/value projections of all attention layers with rank \(8\).  This parameter‑efficient fine‑tuning strategy reduces communication cost and the magnitude of DP noise, particularly beneficial for DP‑FL \cite{ffa_lora, xu2024dp}.

\noindent\textbf{Learning rates:}  
The learning rate is set to \(1\times10^{-2}\) for all GLUE tasks and \(1\times10^{-3}\) for text generation; further optimization would consume additional privacy budget. Recent work \cite{prodigy_opt, bu2025towards} shows that learning‑rate optimization can be performed hyperparameter‑free, but we do not adopt it here to keep the experimental protocol simple.

\noindent\textbf{Public clipping thresholds:}  
For the public clipping thresholds \(\mathbf{s}\) we use the following flattened vector: $\bigl[1.0,\,1.25,\,1.5,\,2.0,\,2.5,$
$\,3.0,\,4.0,\,6.0,\,8.0\bigr]^{\!\top} \times \bigl[10^{-1}, 10^{0}, 10^{1}\bigr]$.
For the multipliers \(\bm{m}\) we use \([0.1,\;0.3,\;0.5,\;0.7,\;0.9,\;1.0]\).

\noindent\textbf{Loss computation:}  
DP‑LAC uses the validation set of the corresponding dataset to compute the server validation loss for the adaptive clipping update described earlier. A client‑loss variant (DP-CLAC) is also evaluated: each client computes the loss on its local training data, privatises it with the Gaussian mechanism (threshold set to the previous round's private mean value), and the resulting noisy loss is sent to the server and mean aggregated.

\noindent\textbf{Methods compared:}
Five baselines are evaluated against DP‑LAC and DP-CLAC: (a) DP‑FedAvg with fixed clipping \cite{dp_sgd}, (b) gradient normalization \cite{adaptive_clipping_bu}, (c) adaptive clipping using quantiles \cite{adaptive_clipping_andrew}, (d) decaying the noise multiplier \cite{adaptive_clipping_du}, and (e) decaying the clipping threshold \cite{ adaptive_clipping_qiu}.  All experiments are conducted using the Flower simulation framework and the Huggingface training library. Additional experiments are conducted to validate the cost, accuracy, and robustness of the privately estimated clipping bound $C_0$.

\vspace{-0.5em}
\section{Results}
\vspace{-0.5em}
\subsection{Time Cost of Hyperparameter Tuning}
Table \ref{tab:grid-search} lists the time cost of tuning hyperparameters for all methods either sequentially (one by one) or exhaustively (all together) in units of $\tau$---the runtime for one complete FL experiment on a given dataset. With 20 parallel workers on 5x 80GB GPUs, $\tau$ is 7.25h, 8.25h and 11h for SST2, QNLI and MNLI respectively. DP-LAC is 5-15x faster compared to prior adaptive and non-adaptive methods with sequential grid search.  We performed sequential grid search for all methods since the time for exhaustive search was infeasible for some methods.

\begin{table}[ht] %bp
  \centering
\resizebox{1\columnwidth}{!}{
    \begin{tabular}{cccc}
    \toprule
    \multirow{2}{*}{\textbf{Method}} & \multirow{2}{*}{\textbf{Hyperparameters}} & \multicolumn{2}{c}{\textbf{Grid Search}}\\
    \cmidrule(lr){3-4} & & \textbf{Sequential} & \textbf{Exhaustive}\\
    \midrule
    Abadi \cite{dp_sgd} & $C$ (clip threshold) & 5$\tau$ & 5$\tau$ \\
    Andrew \cite{adaptive_clipping_andrew} & $\gamma$ (quantile), $\eta_{C}$ (adapt. lr) & 10$\tau$ & 25$\tau$ \\
    Du \cite{adaptive_clipping_du} & $C_0$ (clip threshold), $\rho_{C}, \rho_{\mu}$ (decay rates) & 15$\tau$ & 125$\tau$ \\
    Bu \cite{adaptive_clipping_bu} & $\gamma$ (stability), $\eta$$C$ (lr) & 10$\tau$ & 25$\tau$ \\
    Qiu \cite{adaptive_clipping_qiu} & $\lambda$ (decay rate), $r$ (skip rounds), $\zeta$ (tol.) & 15$\tau$ & 125$\tau$  \\
    \textbf{Ours (DP-LAC)} & - & $\boldsymbol{\tau}$ & $\boldsymbol{\tau}$ \\
    \bottomrule
    \end{tabular}
    }
    \vspace{-0.2em}
  \caption{\label{tab:grid-search}
    The time cost in units of $\tau$ for grid searching 5 choices per DP related hyperparameter sequentially or exhaustively.
  }
\vspace{-1.5em}
\end{table}

\begin{table}[ht] %bp
  \centering
\resizebox{0.85\columnwidth}{!}{
    \begin{tabular}{ccccccccc}
    \toprule
    & & \multicolumn{3}{c}{\textbf{SST2}} & \multicolumn{3}{c}{\textbf{QNLI}} & \multicolumn{1}{c}{\textbf{MNLI}} \\
    \cmidrule(lr){3-5} \cmidrule(lr){6-8} \cmidrule(lr){9-9} %\hline
    \multicolumn{2}{c}{\textbf{Method}} & $\bm{\varepsilon=2}$ & $\bm{\varepsilon=4}$ & $\bm{\varepsilon=8}$ & $\bm{\varepsilon=2}$ & $\bm{\varepsilon=4}$ & $\bm{\varepsilon=8}$ & $\bm{\varepsilon=4}$\\
    \midrule
    \multirow{4}{*}{\rotatebox{90}{Def. HP}} & 
    Abadi (2016) \cite{dp_sgd} & 55.8 & 63.9 & 84.6 & 54.0 & 57.4 & 67.7 & 37.2 \\
    & Andrew (2021) \cite{adaptive_clipping_andrew} & 55.7 & 62.1 & 82.7 & 53.4 & 56.4 & 65.4 & 36.0 \\
    & Bu (2023) \cite{adaptive_clipping_bu} & 55.8 & 63.6 & 84.8 & 54.0 & 57.4 & 67.6 & 37.2 \\
    & \textbf{Ours (w/out $C_\textit{hist}$)} & \textbf{70.5} & \textbf{84.4} & \textbf{89.1} & \textbf{54.0} & \textbf{61.6} & \textbf{70.5} & \textbf{50.1}\\
    \midrule
    \multirow{7}{*}{\rotatebox{90}{DP HPO}} & 
    Abadi (2016) \cite{dp_sgd} & \underline{57.9} & 66.8 & 83.0 & 56.4 & 60.3 & 67.7 & 35.3\\
    & Andrew (2021) \cite{adaptive_clipping_andrew} & 54.9 & 55.3 & 61.5 & 52.1 & 53.5 & 52.3 & 35.2\\
    & Du (2022) \cite{adaptive_clipping_du} & 56.3 & 68.8 & 83.4 & \underline{57.1} & \underline{61.3} & \textbf{68.3} & 43.2\\
    & Bu (2023) \cite{adaptive_clipping_bu} & 57.7 & 66.9 & \underline{83.5} & \underline{57.1} & 60.3 & 67.7 & 42.0\\
    & Qiu (2024) \cite{adaptive_clipping_qiu} & 52.6 & 52.7 & 81.9 & 54.4 & 59.4 & 67.8 & 42.0\\
    & \textbf{Ours (DP-CLAC)} & 55.0 & \underline{73.7} & \textbf{86.4} & 55.6 & 58.0 & 65.1 & \underline{43.6}\\
    & \textbf{Ours (DP-LAC)} & \textbf{67.7} & \textbf{78.1} & \textbf{86.4} & \textbf{59.1} & \textbf{62.5} & \underline{67.9} & \textbf{46.3}\\
    \bottomrule
    \end{tabular}
    }
    \vspace{-0.2em}
  \caption{
    Mean accuracy of 3 runs comparing our method versus previous approaches for DP federated fine-tuning. Under the default hyperparameter selection (Def. HP), our method (w/out $C_{\textit{hist}}$) outperforms all previous approaches. When taking into account the extra privacy cost of DP hyperparameter optimization, our method (DP-LAC) gives either the \textbf{best} or \underline{2nd best} results.}
  \label{tab:dphpo-defhp}
  \vspace{-0.25em}
\end{table}

\vspace{-1.8em}
\subsection{Default and optimized hyperparameters}
For the default hyperparameters scenario (Table \ref{tab:dphpo-defhp} Def. HP), we use $C_0=8$ for all methods including ours instead of $C_0=C_{\textit{hist}}$. Our method is able to beat all baselines for this setting, showcasing the usefulness of adaptively clipping $C$ during training.

We next validate whether tuning hyperparameters under DP are indeed helpful. Table \ref{tab:dphpo-defhp} DP-HPO showcases the effect of DP hyperparameter tuning on several baseline methods. As expected, the DP hyperparameters for the baseline methods evaluated are sensitive to changes in the privacy level $\varepsilon$. For example, all baseline methods under $\varepsilon=2$ generally performed better after hyperparameter tuning even though they had $1/3$ of the original privacy budget.

Under DP-HPO, our proposed methods, DP-LAC and DP-CLAC are able to beat almost all of the evaluated baselines. This demonstrates the efficacy of our method which doesn't require tuning of hyperparameters yet still achieves SOTA performance. We note that DP-CLAC has reduced performance compared to DP-LAC due to a reduced budget available for privatizing model weights.

For the only scenario where DP-LAC doesn't exceed prior art performance (QNLI $\varepsilon=8$), the difference is only $0.6\%$. At this lowest privacy level, it is likely that the gap between $C_{hist}$ and tuned $C$ outweighs the negative affects of increased noise levels due to hyperparameter tuning. Overall, we achieve an average improvement of 6.6\% compared to the previous best across all 3 datasets. 

On the MNLI dataset, a comparatively harder task having the most samples and three output classes instead of two, our method outperforms previous baselines, making it more suitable for realistic scenarios and stronger privacy regimes. We note that the method by \cite{adaptive_clipping_andrew} likely performs significantly poor since they don't consider initial clipping threshold as a hyperparameter.

The lower bounds of the 95\% one‑tailed confidence intervals for DP‑LAC across the three runs are 56.5, 70.1, 84.7, 57.3, 60.4, 66.7, and 44.2. For reference, non-DP FL achieves accuracies of 92.3, 83.2, and 76.6 on SST2, QNLI, and MNLI respectively.

\vspace{-0.6em}
\begin{table}[ht] %bp
  \centering
\resizebox{0.6\columnwidth}{!}{
    \begin{tabular}{cccccc}
    \toprule
     & \multicolumn{2}{c}{\textbf{SST2}} & \multicolumn{2}{c}{\textbf{QNLI}} & \multicolumn{1}{c}{\textbf{MNLI}} \\
     \cmidrule(lr){2-3} \cmidrule(lr){4-5} \cmidrule(lr){6-6} %\hline
    \textbf{Method} & $\bm{\varepsilon=4}$ & $\bm{\varepsilon=8}$ & $\bm{\varepsilon=4}$ & $\bm{\varepsilon=8}$ & $\bm{\varepsilon=4}$\\
    \midrule
    $C_{\textit{hist}}$ & 8.0 & 8.0 & 5.7 & 6.0 & 1.5 \\
    $C_{*}$ & 7.0 & 10.0 & 2.0 & 4.0 & 2.0 \\
    $\textit{Acc}_{\textit{hist}}$ & 78.1 & 84.0 & 62.5 & 66.4 & 46.3 \\
    $\textit{Acc}_{*}$ & 86.8 & 89.4 & 71.5 & 75.6 & 48.4 \\
    \bottomrule
    \end{tabular}
    }
    \vspace{-0.3em}
  \caption{Comparison of how well our one-shot noisy estimate $C_{\textit{hist}}$ tracks the optimally tuned $C_{*}$ based on bias-variance trade-offs. $C_{\textit{hist}}$ is only from clients sampled in the first round. $\textit{Acc}_{\textit{hist}}$ and $\textit{Acc}_{*}$ are accuracies when using $C_{\textit{hist}}$ or $C_{*}$ respectively. \vspace{-1.5em}}
  \label{tab:hist-est}
\end{table}

\begin{table}[ht] %bp
  \centering
\resizebox{\columnwidth}{!}{
    \begin{tabular}{ccccccc}
    \toprule
     & \multicolumn{2}{c}{\textbf{SST2/Rank=4/1B}} & \multicolumn{2}{c}{\textbf{SST2/Rank=16/1B}} & \multicolumn{2}{c}{\textbf{SAMSum/Rank=32/4B}} \\
     \cmidrule(lr){2-3} \cmidrule(lr){4-5} \cmidrule(lr){6-7} %\hline
    \textbf{Method} & $\bm{\varepsilon=4}$ & $\bm{\varepsilon=8}$ & $\bm{\varepsilon=4}$ & $\bm{\varepsilon=8}$ & $\bm{\varepsilon=4}$ & $\bm{\varepsilon=8}$\\
    \midrule
    Bu (2023) \cite{adaptive_clipping_bu} & 75.7 & \textbf{85.7} & 74.1 & 84.4 & 18.1/33.9 & 19.3/35.9 \\
    \textbf{Ours (DP-LAC)} & \textbf{77.9} & 85.0 & \textbf{76.9} & \textbf{85.4} & \textbf{18.9/36.6} & \textbf{20.5/37.8} \\
    \bottomrule
    \end{tabular}
    }
    \vspace{-0.3em}
  \caption{
    Robustness of DP-LAC for different LoRA ranks and tasks. DP-LAC achieves competitive performance under different ranks and exceeds under $\varepsilon=4$ for all ranks. Metrics are accuracy for SST2 and WeightedRouge/RougeL for SAMSUM.
  }
  \label{tab:rank-ablation}
  \vspace{-1.7em}
\end{table}

\subsection{Use of Histogram Statistics}
\label{sec:hist-stats}
\vspace{-0.15em}
Table \ref{tab:hist-est} compares accuracies $Acc_{\textit{hist}}$ and $Acc_{\textit{*}}$ between two choices for the initial clipping threshold, $C_0 = C_{\textit{hist}}$, or $C_0 = C_{*}$, obtained by performing a non-DP hyperparameter search at each privacy budget.  Results show that the DP estimate $C_{\textit{hist}}$ is always within an order of magnitude to the non-private oracle $C_{\textit{*}}$. Thus, histogram estimation can greatly reduces the hyperparameter search space for $C_0$ as compared to all baseline methods evaluated. Additionally, further  performance gains could be achieved via DP-HPO of $C_0$.

\vspace{-0.6em}
\subsection{Effect of changing LoRA Ranks}
\vspace{-0.25em}
Table \ref{tab:rank-ablation} presents the performance of DP-LAC on three additional LoRA ranks. For rank 32, using a higher parameter model (4B) and a text generation task (summarization), DP-LAC is able to beat the baseline \cite{adaptive_clipping_bu} under both $\varepsilon = 4$ and $8$, making it suitable for privately fine-tuning LLMs. It achieves competitive performance under $\varepsilon=8$ for SST2/Rank=4/1B, a comparatively easier task using a smaller model. This is likely because a lower clipping bias is more suitable over a lower noise variance for lower privacy levels. Overall, DP-LAC demonstrates robustness to non-DP hyperparameter changes and can be used as a drop-in replacement to tweak $C$ on every model hyperparameter or dataset change.

\vspace{-0.60em}
\section{Conclusion}
\vspace{-0.75em}
In this paper, we propose DP-LAC, a novel method for dynamically adapting the clipping threshold during the fine-tuning of LLMs in DP-FL, which operates without the introduction of additional hyperparameters. Through empirical evaluations, we show that DP-LAC outperforms state-of-the-art adaptive clipping methods achieving up to 6.6\% improvements on average under the same privacy guarantees. By improving performance and simplifying hyperparameter searches, we hope this work makes privacy-preserving collaborative LLM training more attainable for a broader range of users. In future, we plan on extending our evaluation to other modalities as well as evaluating alternative statistics for estimating the initial clipping threshold.

% % -------------------------------------------------------------------------

\clearpage
\bibliographystyle{IEEEbib}
\bibliography{strings,refs}

\end{document}